# Short-Term Prediction of Demand for Ride-Hailing Services: A Deep Learning Approach

Long Chen[1] · Piyushimita (Vonu) Thakuriah[2] · Konstantinos Ampountolas[3]

## Abstract
As ride-hailing services become increasingly popular, being able to accurately predict demand for such services can help operators efficiently allocate drivers to customers, and reduce idle time, improve traffic congestion, and enhance the passenger experience. This paper proposes UberNet, a deep learning convolutional neural network for short-time prediction of demand for ride-hailing services. Exploiting traditional time series approaches for this problem is challenging due to strong surges and declines in pickups, as well as spatial concentrations of demand. This leads to pickup patterns that are unevenly distributed over time and space. UberNet employs a multivariate framework that utilises a number of temporal and spatial features that have been found in the literature to explain demand for ride-hailing services. Specifically, the proposed model includes two sub-networks that aim to encode the source series of various features and decode the predicting series, respectively. To assess the performance and effectiveness of UberNet, we use 9 months of Uber pickup data in 2014 and 28 spatial and temporal features from New York City. We use a number of features suggested by the transport operations and travel behaviour research areas as being relevant to passenger demand prediction, e.g., weather, temporal factors, socioeconomic and demographics characteristics, as well as travel-to-work, built environment and social factors such as crime level, within a multivariate framework, that leads to operational and policy insights for multiple communities: the ride-hailing operator, passengers, third-part location-based service providers and revenue opportunities to drivers, and transport operators such as road traffic authorities, and public transport agencies. By comparing the performance of UberNet with several other approaches, we show that the prediction quality of the model is highly competitive. Further, UberNet's prediction performance is better when using economic, social and built environment features. This suggests that UberNet is more naturally suited to including complex motivators of travel behavior in making real-time demand predictions for ride-hailing services.

✉ Long Chen
long.chen@glasgow.ac.uk

Piyushimita Vonu Thakuriah
p.thakuriah@rutgers.edu

Konstantinos Ampountolas
k.ampountolas@uth.gr

[1] 7 Lilybank Gardens, Glasgow, UK

[2] Edward J. Bloustein School of Planning and Public Policy, Rutgers University, New Jersey 08901, USA

[3] Department of Mechanical Engineering, University of Thessaly, 38334 Volos, Greece

## Introduction

With the advent of Web 2.0, ride-sharing platforms such as Uber and Lyft are becoming a popular way to travel. For instance, over 14 million trips were provided daily in more than 700 cities around the world by Uber alone[1]. To efficiently manage ride-hailing service operations, there is a need to have fine-grained predictions of the demand for such services, towards the goal of allocating drivers to customers in order to maximize customer service and revenue.

This paper proposes UberNet, a deep learning convolutional neural network (CNN) for the short-term demand prediction of ride-hailing services, using spatio-temporal data of Uber pickups in New York City and exogenous features to

---
[1] https://www.uber.com/en-GB/newsroom/company-info/.

Springer

enhance the learning ability of deep neural networks (DNN). A core indicator of Uber demand is the so-called unit original pickup (UOP), which is the number of successful Uber-transactions at the platform per unit time (Tong et al. 2017). UOP prediction can benefit Uber platform or other ride-hailing services in multiple ways. First, by examining historical UOP data, the platform can find times and geographical areas indicating strong demand. Second, it can be used to assess incentive mechanisms and dynamic pricing policies. UOP reflects the willingness of users to complete transactions after adopting new dynamic pricing policies. Finally, it can be used to optimize ride-hailing services car allocation. This also allows other taxicab platforms to arrange roaming drivers to customers in advance. Predicting UOP accurately is at the core of the online ride-sharing industry.

The basic problem we consider is as follows: let $\{\mathbf{X}(t_0), \mathbf{X}(t_1), \ldots\}$ be a vector time series with elements of a number of temporal and spatial features that have been found in the literature to explain demand for ride-hailing services. Our goal is to predict the count of pickups (Uber or other ride-hailing services) at the future time $t + \delta$ (where the time-interval $\delta$ is given), utilising the proposed deep learning approach UBERNET. Traditionally, short-term prediction problems in transport widely utilise univariate data that are generated from specific sources, e.g., loop detection, GPS or other types of data (Vlahogianni et al. 2004). A comprehensive approach utilising additional geographical, demographic and social and economic factors has been reported in the literature to a lesser degree, with such features being mainly used in travel behavior analysis to understand who, where, and how ride-hailing services are used or what the effects on demand for other transport modes are. Furthermore, transport demand usually follows a diurnal cycle with, for example, a morning and evening peak, and which Uber pickups also exhibit (Chen et al. 2015). In addition, an extensive literature on transport operations has showed the importance of inclement weather, special events and other short-term events on transport use patterns. Our goal with UBERNET is to utilise such additional operational, social and economic features for the short-term prediction problem to capture the range of spatial and other heterogeniety in the Uber demand estimation problem, thereby leading to a multivariate approach to short-term forecasting of ride-hailing demand. We find that the use of such features within the proposed UBERNET approach improves "real-time" short-term demand prediction.

The benefits of having more accurate short-term forecasts of ride-hailing demand are many, and are reviewed in detail in the next section. We have proposed UBERNET as an approach that can formally utilise the multivariate architecture while also being able to utilise the correlations among different features and the long-term dependencies of time series sequences, unlike many other deep learning approaches. The proposed method can effectively capture the time-varying spatio-temporal features of Uber demand at different boroughs. In addition to introducing the prediction approach and assessing the value of multiple features in the prediction problem, the paper compares the performance of the UBERNET with several competing approaches (e.g., ARIMAX, PROPHET, TCN and LSTM) in a number of different settings. The results show that the proposed method is more principle and robust when incorporating and exploiting a wide range of exogenous features such as the four sets of features we utilised in the paper.

Both the popular press and scientific community have addressed many social and economic aspects of ride-hailing companies such as Uber. A significant part of the attention has been negative, in keeping with overall criticisms of new forms of work emerging in the gig economy. Within those overall complex socioeconomic concerns, improvements in short-term demand prediction methodology has several potential benefits. First, the operator may be able to apply surge-pricing more accurately. Other authors have noted that if surge pricing is effectively predicted and disseminated to both drivers and riders, a number of key operational decisions become possible, including the ride-hailing company's ability to more efficiently allocate vehicles, which ultimately helps the efficiency and reliability of transport networks. Other directions in which there are discussions is in spatial pricing, as well as in spatio-temporal bonus pricing as incentives for drivers to offer services in areas with lower levels of vehicle supply.

Second, drivers may be able to reduce idle time and uncertainty about repositioning, thereby reducing lost earnings. However, we mentioned previously, it is not clear how much of the revenue stream arising from information certainty will actually go into the pockets of the drivers. That is an important limitation of the extent to which system accuracy can really benefits drivers.

Third, it is possible that third-party providers of ride-hailing driver assistance may be able to better capitalise on reliable predictions to provide a number of location-based services to drivers, including the potential to earn more. Various incentive mechanisms may also be developed to improve the driver's bottomline, based on different markets served.

Fourth, passengers may be able to reduce wait time if they know that demand for rides is high, thereby either postponing their trip or using other modes of transport. Ride-hailing platforms often use reward systems and promotions to improve customer savings and boosting of both short- and long-term demand. One of many different strategies used is to provide special discounted prices during specified times. It is possible to expand this strategy by incentivizing passengers to defer their trip or travel by other modes of transport at times which are predicted to be busy, with incentive solutions such as mileage programs or the ability to offset a portion of the fares of trips subsequently taken using rewards.

Finally, operators of other transport modes can benefit from knowing about variations in predicted ride-hailing demand, especially in areas, where ride-hailing is popular. For example, road operators can be better prepared through signalization, dynamic ramp metering or other real-time demand management strategies to direct and route traffic due to knowing predicted peaks and troughs in advance. Some cities, including New York and Chicago are now charging congestion pricing of Uber rides. It is possible that better forecasts can help set dynamic congestion pricing to reflect real-time impact of ride-hailing on traffic levels.

The paper is organized as follows: in the next section, we present motivations for the short-term forecasting problem of ride-hailing demand, and deep learning approaches that have been utilised in similar context. Then, in the following section, we describe the methodological approach we are taking to the forecasting problem, by first discussing CNN which helps us set up the UberNet approach. In the next section, we discuss the implementation of UberNet in the context of Uber demand prediction in NYC, including exploratory analytics of features, and setting up of the competing models. Results are presented in the next section and our conclusions in the last section.

## The Ride-Hailing Short-Term Demand Forecasting Problem

In this section, we explore three aspects in the literature that underpins our paper: (1) the uses, impacts and other motivations for accurate short-term forecasts of demand for Uber rides; (2) factors affecting ride-hailing demand that are likely to be important to consider in the forecasting problem; and (3) machine learning methods that have been used in the short-term ride-hailing or similar passenger demand prediction conditions.

*Uses, impacts and motivations for accurate short-term forecasts of demand for Uber rides*: One of the biggest benefits of having more accurate forecasts of where demand exists or where demand will be high in the near future is reduction in vehicle idling time. Vehicle idling has a number of negative consequences. Drivers also tend to behave very differently when demand is not known. For example, if there is place to park, drivers would just stay in one place and wait. Otherwise one may cruise the streets. These are likely to lead to negative environmental impacts from increased traffic congestion and diversion of passengers from public transport (Zhang and Zhang 2018). Yet other drivers, based on historical knowledge of demand patterns preemptively go towards areas, where there is the potential to pick up passengers. One direction of work involves improved operations through algorithms aimed at improving empty vehicle routing and repositioning for both taxi and ridesourcing systems (Iglesias et al. 2018; Yu et al. 2019). Uber uses a proprietary and opaque real-time pricing strategy to equilibrate short-term demand and supply, which is called "surge pricing", leading to variations in prices that range from the base price to prices that are five or more times higher (Cohen et al. 2016; Cramer and Krueger 2016). This approach is estimated to lead to greater efficiency of ride sharing services vis-a-vis taxis. Others have found that Uber partners drive more at times when earnings are high and flexibly adjust to drive more at high surge times (Chen 2016). One goal of the UberNet is to capture the upward or downward spikes in demand resulting from the use of the pricing scheme.

More accurate forecasts have a number of other uses. Knowing likely areas with higher demand in the next timestep allows more accurate setting of dynamic surge pricing. Dynamic pricing is applied by ride-hailing companies for a number of reasons such as weather-related events, special events, and so on, when demand in an area and at a specific time tends to be high. Dynamic pricing serves as an incentive for Uber drivers to drive to areas of the city, where demand for rides is higher, and thereby helps the operator with real-time adjustment and proactive positioning of the ride-hailing fleet to changing demand patterns. However, little is known about the proprietary surge pricing process applied by companies (Battifarano and Qian 2019). These benefits should be contextualized within research literature that has pointed to significant limitations of the "gig economy" model. Extensive previous research has shown that the ride-hailing compensation model poses problems for drivers, because of the risk of not getting work due to demand uncertainty leading to lost wages for drivers as they await new passengers, not knowing, where demand might be high yielding rides which have the potential to give them income. Li et al. (2019) note that drivers in the ride-hailing world who are classified as independent contractors bear the greatest levels of risk due to demand uncertainty. In addition to loss due to demand uncertainty, drivers face the additional problem of low overall compensation after deducting Uber fees and driver vehicle expenses from passenger fares (Zoepf et al. 2018), leading to overall low earnings.

An important outcome of the earnings issue that has received extensive coverage in the popular press is that a large arsenal of data, analytics and technology are now being used to accurately match available drivers to incoming requests leading to improved service availability, shorter waiting times, and ultimately a boost in the company's profits. However, these profits have still not benefited workers and and asymmetries of information and power have persisted (Rosenblat and Stark 2016). A new generation of third-party location-based services (LBS) have sprung up to support drivers of ride-hailing services which purports to help drivers earn more through different types of bundled service options while also providing opportunities for ride locations that are matched to driver preferences. Some examples in the US are Gridwise, Rideshare Guy, PredictHQ and Cargo. These apps target drivers directly by providing a range of information on ride-active areas, in-car commerce and goods which are for sale to passengers, in-vehicle location-based ads and so on. While these systems have their own proprietary sources of information, they would also benefit from more accurate demand prediction models. However, the potential loss of revenue due to dropped rides resulting from long passenger wait times and uncertainty about the wait could be quite significant, and is another motivation for more accurate demand prediction. Passenger demand is also negatively affected by surge pricing (Chen et al. 2015). Although passenger behavior in ride-hailing services has not been as extensively examined as driver behavior, the issues that can arise include cancellation by the passenger because of a long wait before being assigned to a vehicle, passenger cancellation because of a longer-than expected pick-up time after being assigned a vehicle, and passenger reorder and rebooking after cancellation (Wang and Yang 2019).

*Factors affecting Uber demand*: Whether to use UBER is a complex decision and is directly or indirectly affected by several factors. The factors that motivate the use of Uber and other ride-hailing services have been examined by several researchers. For example, Alemi et al. (2018) noted that among millenials and Gen X travelers, greater land use mix and more urban locations are associated with higher adoption of on-demand ride services. Others, for example Young and Farber (2019), found ride-hailing to be "a wealthy younger generation phenomenon". In the New York City (NYC) area, previous research indicated the importance of lower transit access time (TAT), higher length of roadways, lower vehicle ownership, higher income and more job opportunities in explaining higher taxi/Uber demands (Diego Correa and Ozbay 2017), while others have noted bike share infrastructure, greater share of built-up area, and a higher percentage of residential- and retail-based floor area as having an effect on greater demand for Uber service (Gerte et al. 2018).

More generally speaking, a voluminous literature has examined the direct and indirect factors affecting transport demands. Such factors can generally be categorised into those that can vary in the short-term, and others that are more stable and have long-term, indirect effect on travel demand. People's activity patterns, for example, work, or social activities, drive travel demand (Bhat and Koppelman 1999) and can lead to specific hourly, daily, weekly or seasonal patterns in travel demand and resulting congestion levels (Kim and Kwan 2018). Other exogenous factors, such as inclement weather can affect short-term demand differently by mode, as noted in the case of road traffic (Thakuriah and Tilahun 2013), public transport (Tang and Thakuriah 2012), and bicycling (Meng et al. 2016). The significant relationship of inclement weather to taxi use (Chen et al. 2017) and ride-hailing (Brodeur and Nield 2017) has also been pointed out. Other long-term socio-economic factors, for example, income levels or employment activity, can also affect travel demand. Researchers have pointed to the link between land-uses in surrounding areas and travel demand, pointing, for example, to the critical role of density, land-use diversity, and pedestrian-oriented designs (Cervero and Kockelman 1997). Others have pointed to the importance of socioeconomic factors, for example of income, which has critical consequences on car ownership, and the effect on transport mode and other travel demand behaviors (Nguyen et al. 2017). Finally, factors such as prevailing crime and even perception of crime, can affect travel demand in a long-term sense (Tilahun et al. 2016). Our modelling approach recognizes the complexity of the demand process, and incorporates multiple factors that has the potential to affect ride-hailing demand.

*Related machine learning methods*: Short-term prediction of transport demand is a well-researched topic in multiple transport modes. Recent advances in deep learning have provided new opportunities to address short-term transport behavior. Deep learning often involves the training of a neural network which is computationally prohibitive and prone to over-fitting when compared to traditional approaches, see, e.g., Ramdas and Tibshirani (2016). Various deep learning models have been proposed to analyze time series data and have achieved state-of-the-art performance in real-world applications. For example, Huang et al. (2014) presented a deep belief network to identify the spatio-temporal features and proposed a multi-task learning structure to perform road traffic flow prediction. Similarly, Lv et al. (2014) used a stacked autoencoder model based on traffic prediction method. Tan et al. (2016) studied different pre-training approaches of DNN for traffic prediction. Yang et al. (2015) proposed a stack denoise autoencoder method to learn hierarchical topology of urban traffic flow.

Researchers typically employ recurrent neural networks (RNN) and long short-term memory (LSTM) to intelligently capture spatio-temporal features. They pair this with a softmax activation function node to learn the correlations between the time series points. For example, Wang et al. (2018) used LSTM to predict hourly Uber demand in NYC, using univariate pick-up data, and compared the results to Poisson regression and to regression trees. Wang et al. (2019) uses a standard LSTM layer to capture spatial information, which is then fed into multiple Bi-LSTM layers to incorporate temporal information. Wu et al. (2018) used attention model to determine the correlation between the flow of past spatio-temporal positions and the future traffic flow. Ke et al. (2017) proposed a novel convolutional long short-term memory network to identify spatio-temporal features in a unified framework, which is then coordinated in a end-to-end learning structure for the short-term passenger demand prediction. More recently, Zhang et al. (2019) proposed the fusion convolutional long short-term memory network with residual mapping to capture spatio-temporal correlations in passenger demand. These methods, however, all relied a fully-connect structure, where it is often difficult for the neural networks to capture representative features from the dataset with a large number of complex features. Those results show that the the use of spatio-temporal features can often yield superior predictive performance than RNN alone.

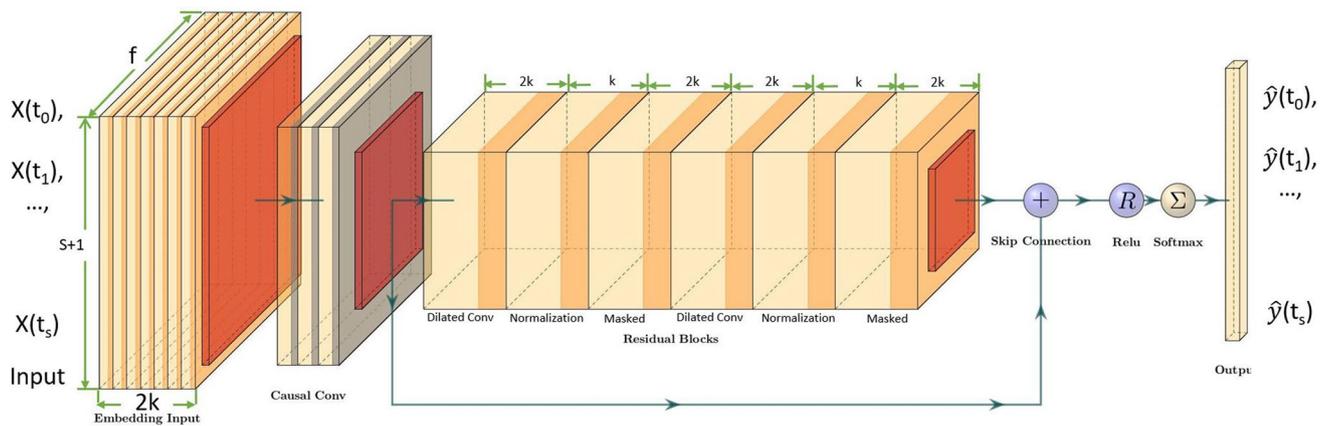

**Fig. 1** Core structure of UberNet

Yang et al. (2019) proposed a hybrid deep learning architecture, where the temporal features along with other traffic-related data are modelled using an LSTM model. The spatial correlations are modeled through a graph based convolutional neural network (GCN). While generally effective, these models, see, e.g., Laptev et al. (2017); Polson and Sokolov (2017), largely rely on the hidden state of the entire past. This makes it hard to apply parallel computing within the series. A number of other NN-based frameworks for spatio-temporal network learning and prediction have been proposed in literature (Zhang et al. 2017; Yao et al. 2018; Wang et al. 2017; Lin et al. 2018; Xu et al. 2018a). These approaches in general offer good performance but require expensive fine-tuning of their parameters for a particular dataset. They either entail a super-linear execution time (Zhang et al. 2017) or convert the original time series into a constant size format (Tong et al. 2017; Xu et al. 2018b), which incurs a memorization step with additional cost.

The WaveNet neural network (Van Den Oord et al. 2016a) has been recently employed to address the aforementioned limitations for time-series prediction and demonstrated its ability to achieve highly competitive performance to the state-of-the-art (Lipton et al. 2015; Gers et al. 2002; Borovykh et al. 2018). A WaveNet-style architecture can be also used to learn local spatio-temporal patterns (Kechyn et al. 2018). The next section develops the proposed UberNet, a deep learning CNN structure. Unlike previous temporal convolutional networks, UberNet employs various temporal and spatial features (e.g., socio-economic variables, spatial heterogeneity, weather, etc.) coupled with raw counts of ride-hailing pickups that are used in a multivariate architecture for the efficient short-time demand prediction of ride-hailing services. The designed CNN architecture and the use of temporal and spatial features have been proved beneficial in enhancing the learning ability of the developed deep neural network.

## UberNet: A Deep Learning Convolutional Neural Network

This section describes UberNet, a deep learning CNN-based approach for the short-term demand prediction of ride-hailing services. The basic architecture of UberNet is based on WaveNet (Van Den Oord et al. 2016a), which was initially proposed for generating raw audio waveforms.

### Architecture

Figure 1 depicts the network architecture of UberNet, including the embedding input, convolutional layers, residual blocks, activation functions, and output. The deep learning CNN architecture of UberNet, takes an input embedding $\mathbf{X} = \{\mathbf{X}(t_0), \mathbf{X}(t_1), \ldots, \mathbf{X}(t_s)\}$ and outputs $\hat{y}(\mathbf{X}) = p(t_{s+\delta})$, where $p(t_{s+\delta})$ is the predicted ride-hailing service pickups (e.g., Uber pickups) at time $t + \delta$, given measurements up to time $s$, via a number of different hidden convolutional layers of abstraction. Here $\{\mathbf{X}(t_0), \mathbf{X}(t_1), \ldots, \mathbf{X}(t_s)\}$ is a vector time series comprising the pickups of ride-hailing services (e.g., Uber) $\{p(t_0), p(t_1), \ldots, p(t_s)\}$ and a number of temporal and spatial features $f$ that have been found to explain demand in ride-hailing services. It should be noted that $f$ can be further divided into two distinctive types, namely space-independent features (e.g., Feature set A in Table 1, see "Deep learning for Uber demand prediction in NYC") and space-dependent ones (e.g., Feature set B, C, and D in Table 1, see "Deep learning for Uber demand prediction in NYC"). Space-independent features can be created by taking the average of all the values within that time interval. Space-dependent features can be created by taking the average of both the time (e.g., 15- and 30-min interval) and the boroughs, so as to enable them to take into account both time and space variations.

UberNet embeds the previous $s$ timestamp as a $(s+1) \times f$ matrix by taking the embedding query operation, where $f$ is the embedding vector size (c.f., the first layer in Fig. 1). Thus, each row of the matrix mapped into the latent features of one timestamp. The embedding matrix is the "cache" of the $s+1$ timestamp in the $f$-dimensional embedding. Intuitively, models of various CNNs that are successfully applied in series transformation can be customized to model the "cache" of a time-dependent traffic demand. However, the demand sequences in real-world entails a large number of "cache" of different sizes, where conventional CNN structures with receptive field filters usually fail. Moreover, the most effective filters for text applications cannot easily fit into modelling demand sequence "caches", since these filters (with respect to row-wise orientation) often fail to learn the representations of full-width representation (see "Dilated convolution").

Given the normalised time series, we propose to use filters (Asiler and Yazıcı 2017), that traverse the full columns of the sequence "cache" by a single large filter. Specifically, the width of filters is equal to the width of the input "cache". The height typically varies with respect to the sliding windows over timestamp at a time. To this end, UberNet is operated directly on a vector time series sequence $\mathbf{X} = \{\mathbf{X}_1, \ldots, \mathbf{X}_{s+1}\}$. The joint probability $P(\mathbf{X})$ of the waveform $\mathbf{X}$ is given as the form of conditional probabilities as follows:

$$\pi(\mathbf{x}) = \prod_{t=1}^{s+1} P(\mathbf{X}_t \mid \mathbf{X}_1, \ldots, \mathbf{X}_{t-1}), \tag{1}$$

where $P(\mathbf{X}_t \mid \mathbf{X}_1, \ldots, \mathbf{X}_{t-1})$ are conditional probabilities. Each datapoint $\mathbf{X}_t$ is, therefore, conditioned on the value of all previous timesteps. Here, every conditional distribution is modelled by a stack of convolutional layers (see Fig. 1). To learn the conditional distributions $P(\mathbf{X}_t \mid \mathbf{X}_1, \ldots, \mathbf{X}_{t-1})$ over the individual timesteps, a mixture density network or mixture of conditional Gaussian scale mixtures can be employed (Theis and Bethge 2015). However, a softmax distribution can give a superior performance to the final layer, even if the data is only partially continuous (Asiler and Yazıcı 2017) (as is the case for special events or holidays in Uber demand data).

Unlike traditional CNN structure that considers the input matrix as a 2D "cache" during convolution operation, UberNet stacks the "cache" together by mapping a look-up table. As has been evidenced in Kim and Kwan (2018), the optimal results is achieved when the number of embedding is set as $2k$, where $k$ is the size of inner channel of the CNN network. In addition, to better capture the spatio-temporal interactions from the 2D embedding input, we conduct a reshape operation, which acts as a prerequisite for operating the 1D convolution in UberNet. We have one dilated filter of size $1 \times 3$ and two regular filters of size $1 \times 1$. The $1 \times 1$ filters are introduced to change the size of channel which can reduce the parameters to be learned by the $1 \times 3$ kernel. The first $1 \times 1$ filter is to change the size of channel from $2k$ to $k$, while the $1 \times 1$ filters does the opposite transformation to maintain the spatial dimensions for the next stacking operation (see the residual blocks in Fig. 1). Since filters of different length will lead to variable-length feature map, max pooling operation is performed over each cache, which selects only the largest value of it, resulting into a $1 \times 1$ cache feedforward layers. The cache from these filters are concatenated to form a feature embedding, which is then fed into a softmax layer (see last layer in Fig. 1) that yields the probabilities of next timestamp.

The core building block of the UberNet is the dilated causal convolution layer (see Fig. 1), which exploits some key techniques such as gated activations and skip connections (see below). In the sequel, we first explain this type of convolution (causal and dilated) and then we provide details on how to implement the residual layers through skip connection. Causal convolution (see "Causal convolution") is employed to handle temporal data, while a dilated convolution (see "Dilated convolution") is used to properly handle long-term dependencies.

### Causal Convolution

In a traditional 1-dimensional convolution layer, we slide a receptive field of weights across an input series, which is then applied to the overlapping regions of the series. Let us assume that $\hat{y}_0, \hat{y}_1, \ldots, \hat{y}_s$ is the output predicted at time steps that follow the input series values $\mathbf{X}(t_0), \mathbf{X}(t_1), \ldots, \mathbf{X}(t_s)$. Since $\mathbf{X}(t_1)$ influences the output $\hat{y}_0$, we use the future time series to predict the past, which will cause serious problems. Using the future data to influence the interpretation of the past one seems to make sense in the context of text classification, since later sentences can still influence the previous ones. In the context of time series, we must generate future values in a sequential manner. To address this problem, the convolution is designed to explicitly prohibit the future from influencing the past. The inputs can only be connected to the future time step outputs in a causal structure. In practice, this causal 1D structure is easy to implement by shifting traditional convolutional outputs by a number of timesteps.

**Dilated Convolution**

One way to handle long-term dependencies is to add one additional layer per time step to reach farther back in the series (to increase the output's receptive field) (Neville and Jensen 2000). With a time series that extends over a year, using simple causal convolutions to learn from the entire history would exponentially increase the computational and statistical complexity. In UBERNET, instead of employing standard convolutions, we designed the dilated convolution to create the generative model, where the dilated layer acts as the convolutional filter to a field which is broader than its original area via the dilation of a zeros sparse matrix. This operation allows the model to become more efficient as it requires fewer parameters. Another advantage is that dilated layer does not change the spatial dimensions of the input, so that the stacking will be much faster on the convolutional layers and the residual structures. Formally, for a vector time series $\mathbf{X}(t_0), \ldots, \mathbf{X}(t_l)$ and a filter $f : \{0, \ldots, k-1\} \to \mathbb{R}$, the dilated function $F$ on element $t$ is given as

$$F(t) = \sum_{i=0}^{k-1} f(i) \cdot \mathbf{X}(t - d \cdot i) \tag{2}$$

where $d$ is the dilation factor, $k$ is the filter size, and $t - d \cdot i$ represents the direction of the past. Dilation thus works as a fixed step between every two adjacent filter taps. If $d = 1$, a dilated convolution boils down to a regular convolution.

The dilated convolutional operation is more powerful to model long-range time series, and thus does not require the use of large filters or additional layers. Practically speaking, one needs to carry out the structure of Fig. 1 multiple times by stacking to further improve the model's capacity. In addition, we employ a residual network to wrap convolutional layers by a residual block, so as to ease the optimization of the deep neural network.

**Masked Residual**

The logic of residual learning is that several convolutional layers can be stacked as a block (see the residual blocks layer in Fig. 1). This allows multiple blocks communicate with each other through the skip connection scheme by passing signature feature of each block. The skip connection scheme can directly train the residual mapping instead of the conventional identity mapping scheme. This scheme not only maintains the input information but also increased the values of the propagated gradients, resolving the gradient vanishing issue. A residual block comprises of a branch leading out to several transformations $\tau$, the outputs of which are forwarded to the input $x$ of the block:

$$o = \text{Activation}(x + \tau(x)) \tag{3}$$

These operations allow the layer to learn the modifications of the identity mapping rather than the entire transformation, which has been known to be useful in deep learning networks.

UBERNET employs two residual modules, as shown in Fig. 1 (Niepert et al. 2016). Each dilated convolutional layer is encapsulated into a residual. The input layer and convolutional one are stacked through a skip connection (i.e., the identity line in Fig. 1). Each block is represented as a pipeline structure of several layers, i.e., normalization layer, activation layer, convolutional layer, and a softmax connection in a specific manner. In this work we put the state-of-the-art normalization layer before each activation layer, which has shown superior performance than batch normalization when it comes to sequence processing. The residual connection allows each block's input to bypass the convolution stage and then adds that input to the convolution output.

**Activation Function**

UBERNET employs the following gated activation unit in the residual blocks when conducting the stacking operation among multiple residual blocks:

$$\mathbf{z} = \tanh(\mathbf{W}_{f,k} * \mathbf{X}) \odot \sigma(\mathbf{W}_{g,k} * \mathbf{X}), \tag{4}$$

where $*$ is the convolution operator, $\odot$ is an element-wise multiplication, $\sigma(\cdot)$ represents a nonlinear sigmoid function, $k$ denotes the layer index, $f$ and $g$ are filter and gate, and $\mathbf{W}$ is a learnable convolution filter (Van den Oord et al. 2016b). The non-linearity and gated activation unit in (4) can outperform a rectified linear activation function, $\max\{x, 0\}$, as shown in Van den Oord et al. (2016b).

**Training of the Neural Network**

The neural network weights can be trained using deterministic or stochastic gradient descent with an ultimate goal to reduce the root mean square error (RMSE). Alternatively, the neural network outputs can be optimized by maximizing the log-likelihood of the input data with respect to the weights. To overcome overfitting, i.e., weights of large values due to noisy data that can make the neural network unstable, we employ $\mathcal{L}_2$ regularization (weight decay). The cost function under optimization can be expressed as

$$E(\mathbf{w}) = \frac{1}{T} \sum_{t=1}^{T} \left(x_t - \hat{y}_t(x_t)\right)^2 + \frac{\lambda}{2} ||w||^2, \quad (5)$$

where $T$ is the number of training data sets, $\mathbf{w} \in \mathbb{R}^q$ is a $q$-dimensional embedding of weights, $\lambda \in \mathbb{R}_{\geq 0}$ is a regularization (or penalty) term, $\hat{y}_t(x_t)$ denotes the forecast (output) of $x_t$ using input data $x_1, \ldots, x_{t-1}$. Intuitively, if $\lambda = 0$ then the regularization term becomes zero and the cost function represents just the RMSE, and $\lambda > 0$ ensures that $w$ will not grow too large. Equation 5 is optimized using deterministic gradient descent and leads to a choice of weights that strike a balance between overfitting and underfitting the training data. The $\mathcal{L}_2$ regularization can ensure an appropriate value range of weights so that it performs better on unobserved data. Note that $\mathcal{L}_2$ regularization can be combined with $\mathcal{L}_1$ regularization for better results.

The matrix in the last layer of the convolution structure (c.f., Fig. 1 ) has the same size as of the input embedding. The output should be a matrix that contains probability distributions of all timestamps in the output series, where the probability distribution is the expected one that actually generates the prediction results. For practical neural network with tens of millions of timestamps, the negative sampling strategy can be applied to avoid the calculation of the full softmax distributions. Once the sampling size are properly tuned, the performance operated by these negative sampling strategies is almost the same as the full softmax method (Kaji and Kobayashi 2017).

## Deep Learning for Uber Demand Prediction in NYC

The Uber raw data is derived from NYC Taxi and Limousine Commission (TLC),[2] which contains over 4.5 million Uber pickups in New York City. It was previously used in a number of studies (Kamga et al. 2015; Wang et al. 2018; Faghih et al. 2019). The trip information includes day of trip, time of trip, pickup location, and driver's for-hire license number. We have chosen datasets which (a) have been made publicly available, so as to enable other researchers to reproduce our results, and (b) have key characteristics covering a large part of the design space (e.g., day of trip, pickup locations) (Table 1).

The final rectangular panel dataset analysed is in the following format: each unit (the rows) is a time interval $t$ (of 15-min or 30-min aggregations), where the feature $p$ is the count of Uber pickups during a time interval, and remaining features (the predictors) described in Table 1 are: [A] temporal and real-time features (e.g., time-of-day, day-of-week, and hourly weather conditions from Meteoblue[3]) linked by time $t$, [B] averages of demographic and socio-economic features such as income levels and unemployment of census tracks, where pick-ups occurred during the time interval, from the American Community Survey 2008–2011 5-year sample[4] [C] average travel-to-work characteristics such as the proportion of commuters who walk or take public transit in the census tracks in which the pickups are located during the interval, also from the census data and [D] social and built environment characteristics of the census tract, where the pickup occurred during the period, for example of crime levels from the NYC police department,[5] and data on the built environment (e.g., density of transport facilities such as the number of transport stops, stations and other facilities) from the NYC Planning Department.[6]

Figure 2a shows the spatial heterogeneity of demand in NYC on 13/04/2014. Most transactions occur in Manhattan, while Bronx seems to be the borough with the least demand. Specific points with high levels of pickups are the airports (John F. Kennedy (JFK) International Airport and LaGuardia Airport). In terms of a diurnal pattern, Fig. 2b shows the average normalized Uber demand variation over 24 h of weekdays and weekends for the year 2014. The solid lines represent weekdays and dash lines denote the weekend. As can be seen, the highest pickup hour over weekdays is at 1:00 am right after midnight and at 5:00 pm when people commuting back to home after work. It is also interesting to observe that in the weekend, the highest levels of pickup is at 10:00 pm to people going out for recreational activities and at 5:00 am when people return home. Figure 2c shows the average normalized Uber demand variation over the seven days of the week for the year 2014. It is clear that demands are highest on Thursdays, Fridays, and Saturdays. Pickups are lower on Sunday and Monday evenings. In addition, weekdays tend to follow a similar pattern, while weekend shows disparage patterns, which implies that weekdays provide more useful patterns than weekend. This is understandable as people usually have irregular work and sleep styles over the weekend.

### Deep Learning Approaches for Assessment

To assess the performance of the proposed deep learning approach for Uber demand prediction in NYC, the following five approaches were compared:

---

[2] https://www1.nyc.gov/site/tlc/index.page.

[3] https://www.meteoblue.com/

[4] https://www.census.gov/programs-surveys/acs.

[5] https://www1.nyc.gov/site/nypd/stats/crime-statistics/historical.page.

[6] https://www1.nyc.gov/site/planning/data-maps/open-data/dwn-selfac.page.

**Table 1** Features employed in UBERNET for Uber demand prediction in NYC

| Feature | Description |
| --- | --- |
| pickups $p(t_i)$ | Count of Uber pickups in a time interval $t_i$ |
| [A] Temporal and real-time factors | |
| hour | Hour of the day of pickup |
| wed | Weekend (E) or weekdays (D) |
| day | Days of a week |
| month | Month of the year |
| vsb | Visibility in miles to nearest tenth |
| temp | Temperature in Fahrenheit |
| dewp | Dew point in Fahrenheit |
| hd | Hour of the day |
| spd | Wind speed in miles/hour |
| slp | Sea level pressure |
| pcp01 | 1-h liquid precipitation |
| pcp06 | 6-h liquid precipitation |
| pcp24 | 24-h liquid precipitation |
| sd | Snow depth in inches. |
| [B] Pickup area demographic characteristics at census tract level | |
| Unemployment | Umemployment rate |
| Income | Median household income |
| Poverty | Poverty rate |
| Self-employed | Count of population who are self-employed |
| TotalPop | Total population |
| [C] Pickup area resident work and travel-to-work characteristics (census tract level) | |
| Walk | Count of population walking to work |
| Transit | Commuting on public transport |
| Carpool | Count of carpools |
| WorkAtHome | Percentage of people who work at home |
| MeanCommute | Average commute time |
| [D] Pickup area location, social and built-environment (census tract) | |
| streetcrime | Count of street crimes in 2014 |
| borough | Five boroughs of NYC |
| PUMA | Public use microdata areas |
| transp | Count of transport facilities (e.g., transit stops, stations) |

1. Autoregressive integrated moving average with explanatory variables (ARIMAX (Cools et al. 2009)) is used as baseline approach. In ARIMAX the "evolving multivariate" of interest are regressed on its own lagged (i.e., prior) values and the "regression error" is a linear combination of error terms whose values occurred contemporaneously and at various times in the past.
2. The off-the-shelf model PROPHET (Drucker et al. 1997; Taylor and Letham 2018). PROPHET is an additive regression model with a piecewise linear kernel. It automatically detects changes in trends by selecting changepoints from the data.
3. The off-the-shelf model support vector machine, which is based on the classic libsvm. (Drucker et al. 1997).
4. The off-the-shelf model random forest regression (Liaw et al. 2002), which is a meta estimator that fits a number of classifying decision trees on various sub-samples of the dataset and uses averaging to improve the predictive accuracy and control over-fitting.
5. The state-of-the-art long short term memory (LSTM) model that is similar to Wang et al. (2018).
6. The state-of-the-art hybrid CNN-LSTM model that is similar to ?
7. The CNN model that is similar to Van Den Oord et al. (2016a) with extensions and modifications of multivariate settings (Yazdanbakhsh and Dick 2019).
8. The proposed deep learning UBERNET, see Sect. UBERNET.

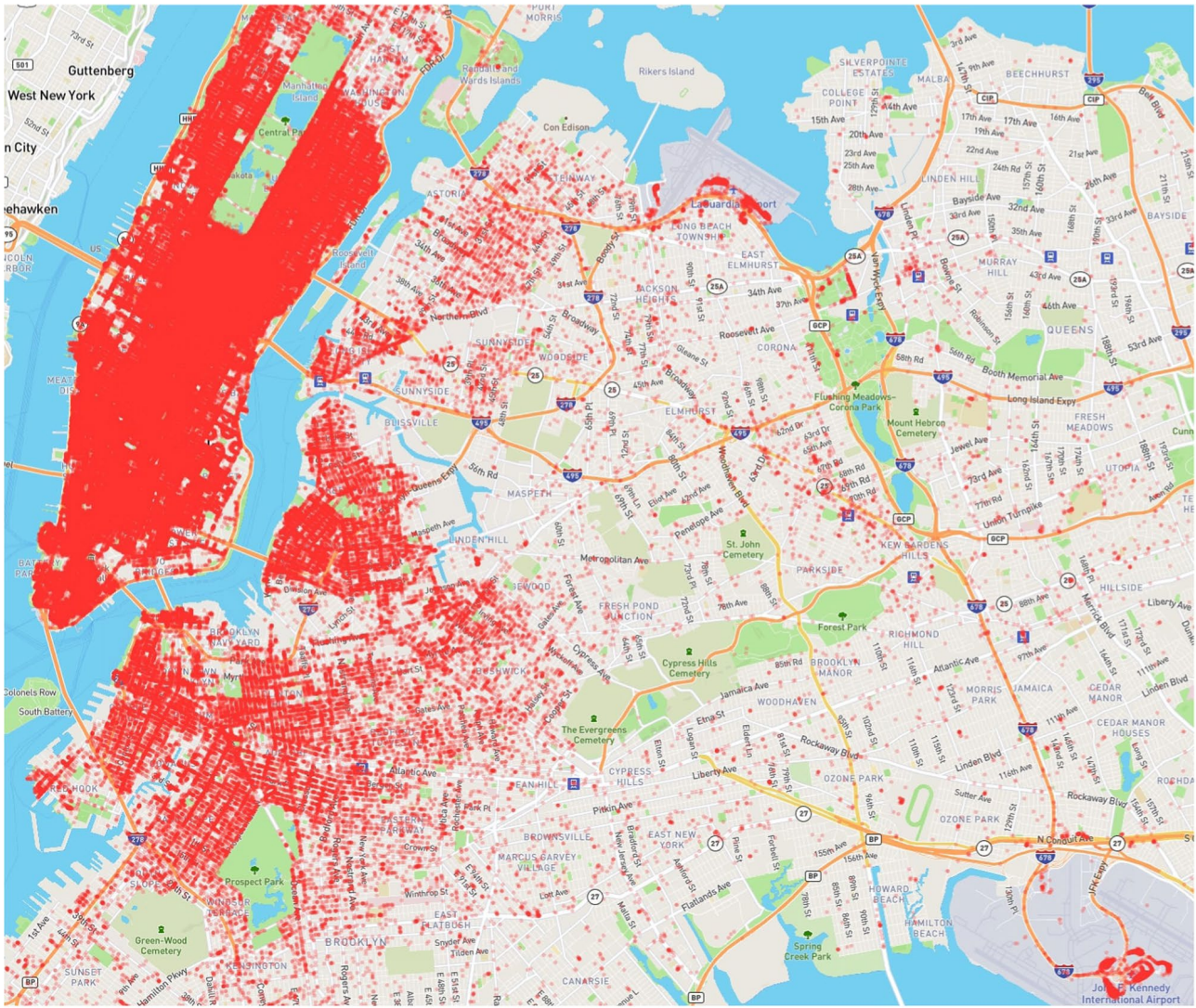

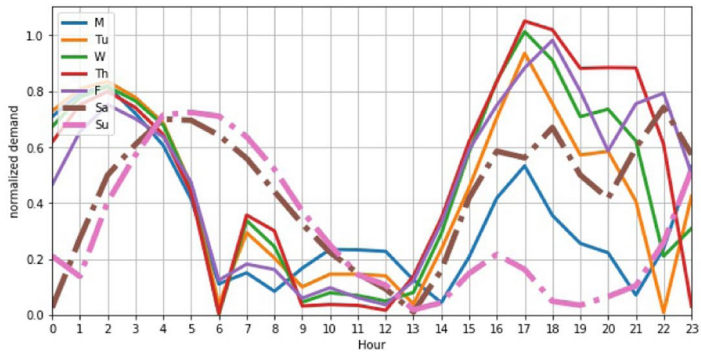

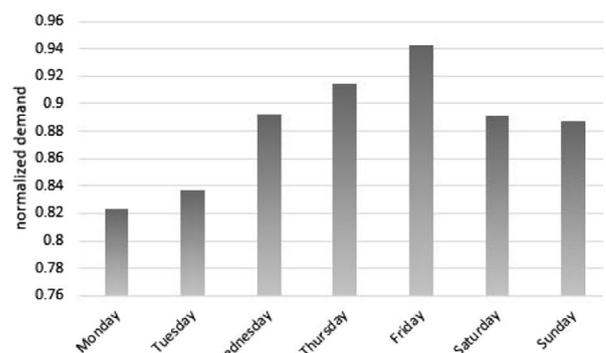

**Fig. 2** **a** Spatial heterogeneity of Uber demand in NYC on 13/04/2014; **b** Uber pickup demands over weekdays; **c** pickup demands over the days of a week

The next section presents the tuned parameters for the above approaches.

## Parameter Settings

For LSTM, CNN-LSTM, SVR, random forest regression, and UBERNET, we apply the grid search (Bergstra and Bengio 2012) on the validation dataset to determine the optimal settings of hyperparameters on the training dataset. These include the latent dimensions, regularization, hyperparameters, the learning rate, and the number of training iterations.

For the ARIMAX model, we employed the adjusted Dickey-Fuller test (Agiakloglou and Newbold 1992), where the model achieved the best performance with $\hat{p} = 2$, $\hat{d} = 1$, and $\hat{q} = 1$. Here, $\hat{p}$ is the number of auto-regressive term, $\hat{d}$ is the number of nonseasonal differences needed for stationarity, and $\hat{q}$ is the number of lagged forecast errors in the prediction.

For the PROPHET model, we set the trend as logistic regression. The monthly seasonal component is modelled using Fourier series, and a weekly seasonal component is captured using dummy variables. The default value of Fourier terms ft (ft = 10) are taken for monthly seasonality. In addition, the strength of holiday model is set as 0.1. The yearly seasonality component is disabled.

A neural network with eight hidden layers (normalization and Relu) with the same number of neurons as there are inputs (28 features, see Table 1) was used for the LSTM model. A rectifier activation function is used for the neurons of 8 hidden layers. A softmax activation function is used on the output layer to turn the outputs into probability-like values. Logarithmic loss is used as the loss function and the efficient ADAM gradient descent algorithm (Ruder 2016) is used to learn the weights. The size of the max pooling windows and stride length are both set to 2. In addition, we set the number of iterations $n$, to 100, the embedding dimension, $f$ to 200, and the learning rate to $10^{-2}$.

In CNN-LSTM, the CNN component has 2 stacked convolutional layers where there are 32 filters of size 2 and 4. There are 6 LSTM layers, where the size of memory cell is set to 600. The number of neurons in the feature fusion layer is set to 900. Furthermore, we adopt dropout and $L_2$ regularization (ridge regression) to prevent overfitting. In addition, the learning rate is set to $10^{-2}$.

As for the SVM, we set the following parameters: the error penalty, $\epsilon$, to $10^{-4}$, the degree of kernel function, $d$, to 3, and regularization parameter, $C$, to 1. With respect to the Random Forest Regression, we set the number of tree to 100. Mean Absolute Error is chosen as the pruning criterion. The minimum number of sample for each tree is set to 1. The aforementioned parameter values found to provide the best results for the random forest.

For UBERNET, we set the number of iterations, $n$ to 100, the embedding dimension, $f$ to 200, and the learning rate to $10^{-3}$. We use teacher forcing (Lamb et al. 2016) during training to solve the problems of slow convergence and instability. This will help UBERNET learn how to condition on the results of the previous timesteps to make a prediction of the next one. At prediction time, we employ a separate function which executes an inference loop to make predictions on test data, iteratively filling the previous $t - 1$ predictions in the history sequence. The coefficients of the convolutional filters are trained by employing an iterative method which alternates between feedforward and backpropagation passes of the training data (Ruder 2016).

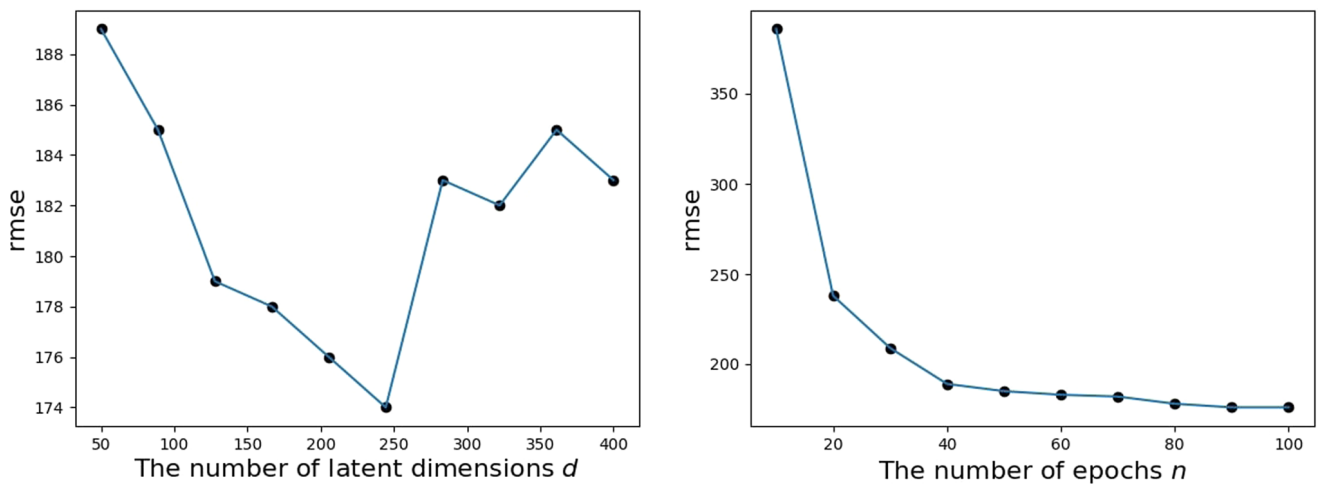

**Fig. 3** Performance of varying the number of latent dimensions $d$ and iteration $n$

**Table 2** Performance of rolling cross-validation over 15-min interval from 20/08/2014 to 30/09/2014

| Metrics | ARIMAX | LSTM | CNN-LSTM | Random forest | SVR | PROPHET | CNN | UBERNET |
|---|---|---|---|---|---|---|---|---|
| SMAPE | 8.24 | 7.53 | 7.48 | 8.19 | 8.23 | 8.24 | 7.79 | 7.31 |
| RMSE | 207.25 | 179.21 | 179.82 | 199.87 | 201.22 | 198.36 | 183.03 | 177.84 |

To assess the quality of fit, we primarily use root mean square error (RMSE) and symmetric mean absolute percentage error (SMAPE), defined, respectively, as

$$\text{RMSE} = \sqrt{\frac{\sum_{t=1}^{n}(F_t - A_t)^2}{n}} \quad (6)$$

$$\text{SMAPE} = \frac{1}{n}\sum_{t=1}^{n}\frac{|F_t - A_t|}{1/2(F_t + A_t)} \quad (7)$$

where $F_t$ is the model's predicted values at timestamp $t$, $A_t$ is the actual value, and $n$ is the number of all the timestamps. RMSE is the classic metric for capturing the general prediction variance, whereas SMAPE is more commonly used in the time series analysis community.

Figure 3 shows RMSE for varying dimension $d$ while keeping all the other parameters with the optimal value. A larger dimension does not always result in a better performance. The best performance is achieved when dimension is properly learned and the mediocre performance of larger dimension may due to over-fitting. Note that for UBERNET, we have fewer total parameters to train than the primitive LSTM architecture, and the model converges with significantly fewer epoch. The model achieves a good performance when dimension falls between 200 and 300. We found that the model tends to give a similar performance when the number of epochs is larger than 100, and thus we use the setting of $d = 200$ and $n = 100$ for the rest of the analysis.

Finally, we used the data between 01/01/2014 and 20/08/2014 for training and the data between the 20/08/2014 to 30/09/2014 for testing.

## Results

The following results are reported with their optimal hyper-parameter settings. Table 2 reports the RMSE and SMAPE for all the 8 approaches for a period of over a month. Cross validation (CV) can usually lead to a more robust result. However, the standard CV leaks future data in the training set. To properly handle time-series, we adopt the 5-fold rolling Cross-validation approach, which is a common practice for time-dependent data in general. Rolling CV is better for time dependent datasets by always testing on data which is newer than the training data.

Table 2 shows that UBERNET performs better than the competitor approaches using both metrics. Based on the RMSE, the table shows that the average prediction error per 15-min time interval is 177.84 pickups, compared to an average prediction error of 207.25 pickups by ARIMAX, the worst-performing method. Over the course of a day, this can hypothetically translate to over 3200 correct predictions by UBERNET compared to ARIMAX. The average NYC UberX ride is estimated to cost approximately $29.34 in 2015,[7] which would translate to potential revenue of over $93,899 per day when compared to ARIMAX if demand were to remain constant; and $12,385.82 per day when compared to CNN if given the same hypothetical condition. The greater accuracy of the UBERNET has the potential to benefit a number of stakeholders in the ride-hailing sector. Generally speaking, LSTM and CNN demonstrated competitive performance, while PROPHET, ARIMAX, SVR and Random Forest performed rather poorly. ARIMAX is the worst due to its shallow structure, while the rest three display a comparable performance. PROPHET is slightly better as it can capture seasonality with a strong robust to outliers. SVR and Random Forest are both strong regression baselines, however, in the context of time series analysis the order of the data points are largely ignored, which lead to compromised results. LSTM is the closest competitor to UBERNET on the basis of both RMSE and SMAPE. UBERNET performs better than the basic CNN indicating that the changes incorporated in "UberNet: a deep learning convolutional neural network" helped with performance gain.

Figure 4 shows all 5 models for 15-min and 30-min look-ahead periods over 10-days. For the 15-min scenario, all the models, except for ARIMAX, do reasonably well in predicting the trough (i.e., the lowest levels of demand, and the levels down to and up from it). The biggest spike in Uber demand was on the evening peak on Aug 22, 2014, which was predicted to a better degree by LSTM and CNN; however, the evening peaks with lower spikes on the two days before, and on several days after are predicted more accurately by UBERNET. Performance generally declines for all models with the 30-min look-ahead, but UBERNET closely tracks the high afternoon spikes. LSTM tracks the huge surge on Aug 22 slightly better than UBERNET but consistently overshoots on the remaining days.

---

[7] https://money.com/uber-lyft-price-per-trip/.

## Feature Importance

In this section, we report the feature importance listed in Table 1. Table 3 reports the performance of pickups prediction with varying feature set. UBERNET achieved the best performance with respect to feature set A (temporal and real-time factors) and feature set D (pickup area location, and social and built-environment factors), which implies that UBERNET is good at mining spatio-temporal patterns. LSTM outperforms all the other models in terms of feature set B (pickup area demographic characteristics), implying that it is possible that an RNN-type structure is more robust to features with a higher percentage of missing values (missing values were imputed by replacing with values of the nearest census tract). PROPHET exceeds all other models when using feature set C (pickup area resident work and travel-to-work characteristics). It is possible that these type of features are noisy and susceptible to overfitting, thereby indicating that simpler models may lead to a better fit to the feature space (Table 4).

To obtain a better understanding of UBERNET's performance with respect to the different features employed, we carry out an ablation analysis on each feature one-by-one. In this set of trials, features are removed from the full feature set one-by-one and then the performance is measured. Intuitively, removing an important feature would lead to a significant performance drop. As expected, pickups is the most important feature as it is the target feature. One can also observe that all features make some contribution to the prediction as none of them can outperform the model with the full feature set. In addition, temporal features, e.g., `hour`, `wed`, and `day` and location features, e.g., `borough`, `income`, and `unemployment` play more important role in the prediction, which is consistent to the previous results reported in Table 3.

We then examine some interesting patterns of exogenous features employed in UBERNET, which is similar to the concept of partial dependence plot in machine learning (Goldstein et al. 2015). Figure 5 shows the predicted pickups against `temp` (temperature levels during pick-up time), and `streetcrime` (street crimes rates), carpool (level of carpool commuters) and `transp` (or density of transport facilities in pick-up area). Here, we follow the same set-up as in "Parameter settings" to produce partial dependence plots, where y-axis represents the predicted pickups generated from UBERNET, and x-axis denotes the mean value of each timestamp (with the 15-min interval) of the selected features, i.e., `temp`, `streetcrime`, `carpool` and `transp`, respectively. The patterns are extremely complex and exhibit nonlinearities, bimodalities, the presence of significant outliers, and many other issues that make short-term predictions difficult. Broadly speaking, Uber demand rise with an increase in temperature but then decline when temperatures become very high. A similar pattern can be seen for street crimes—pickups increase as the level of crime, but only up to a point, after which pickups decline, potentially due to the interaction between crime rates, income levels and dependence on public transport rather than ride-hailing. The relationship of Uber pickups to carpool use is interesting—there is a steep rise in ride-hailing as caropooling levels increase, after which there is a steep drop. Finally, pickups have a bimodal relationship with the density of transport stops, stations, and other facilities in an area, where possibly trip connections can be made—for some type of facilities, ride-hailing remains flat despite increase in transport density, whereas in other cases, pickups rise almost linearly with increase in transport density.

## Spatio-temporal Analysis

Considering the real-world scenarios of UBERNET, another interesting research question is where Uber demand is the hardest or easiest to predict and at what times.

Figure 6a shows the performance of UBERNET over varying hours of a day. The best prediction occurred between 1:00 pm and 8:00 pm when people are carrying out most of their daily activities. The worst prediction hours appear to be between 2:00 am to 7:00 am. One can also observe from Figure 6b, c that the model achieves similar results under these two types of metrics, namely, Manhattan and Queens has a better performance than Brooklyn and Bronx. This is consistent with the observation from the spatial heterogeneity (Drucker et al. 1997; Fig. 2a). Perhaps the superior performance over these two boroughs is due to the fact that the Uber demand is higher in these two areas, which in turn produces more effective signals. Table 5 shows the performance of pickups prediction of varying models over Manhattan and Queens. The result of these two boroughs is slightly better than the results of Table 2, which is a little bit counter-intuitive, since more data usually leads to a better performance. After examining individual boroughs, we find that local models can occasionally outperform global model, since the feature set is tailored to that local region (given enough training instances). However, the performance of the other boroughs are all much worse than Table 2 due to the lesser training data. In addition, as expected, deep learning models outperform all the other types of models, since the non-linearity structure manages to identify and exploit more sophisticated spatio-temporal features. More importantly, UBERNET consistently outperforms all the other models, irrespective of the locations. This suggests that our proposed structure is more effective in capturing the hidden spatio-temporal features from multiple intertwined time series.

We now consider the model performance at different geographical levels. Figure 7 shows the relative feature

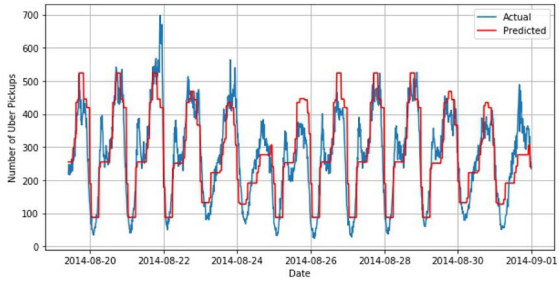
(a) ARIMAX; 15min

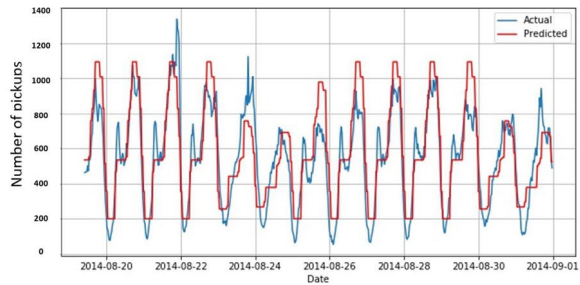
(f) ARIMAX; 30min

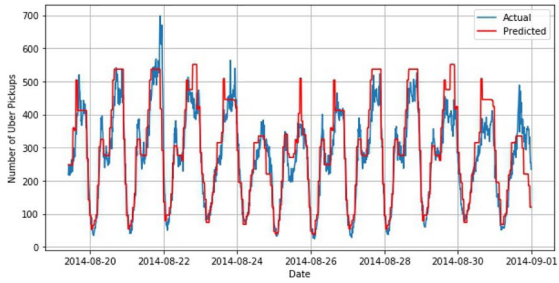
(b) PROPHET; 15min

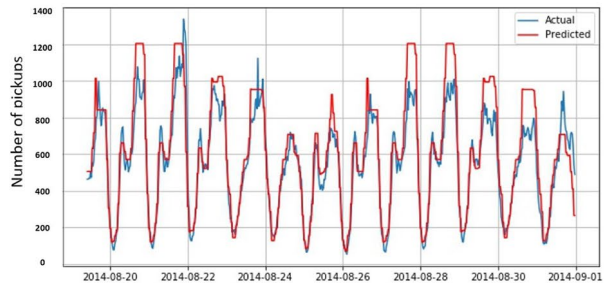
(g) PROPHET; 30min

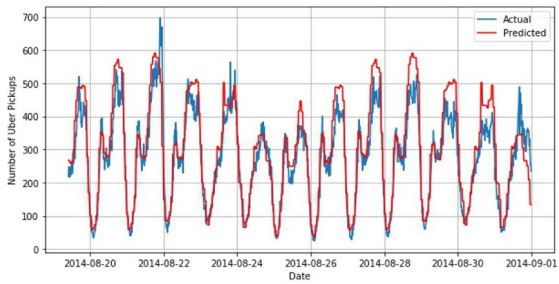
(c) LSTM; 15min

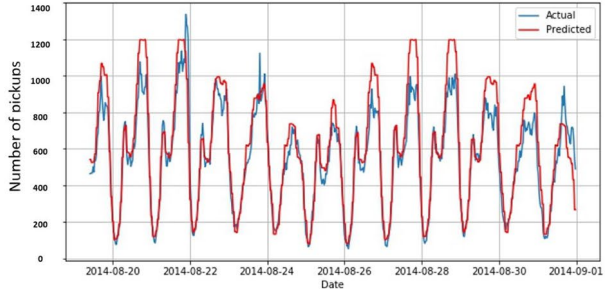
(h) LSTM; 30min

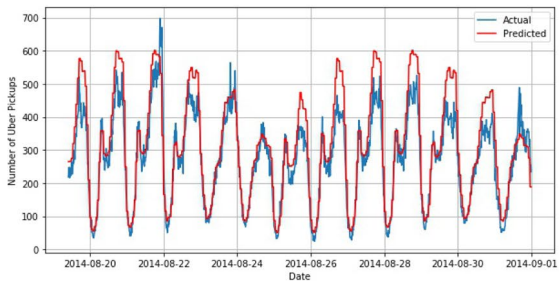
(d) CNN; 15min

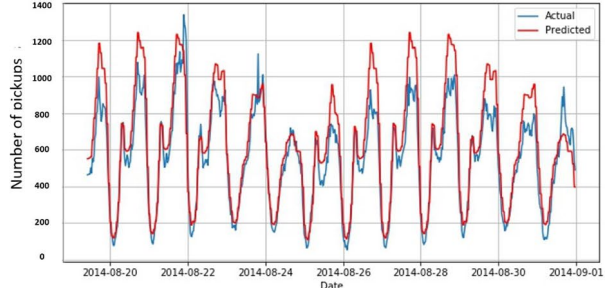
(i) CNN; 30min

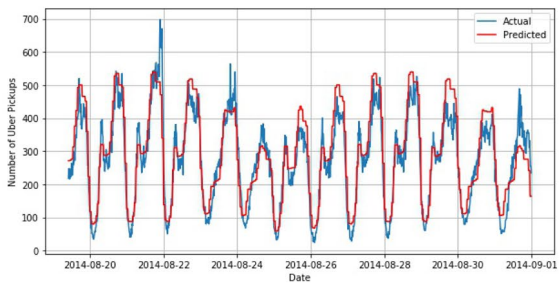
(e) UBERNET; 15min

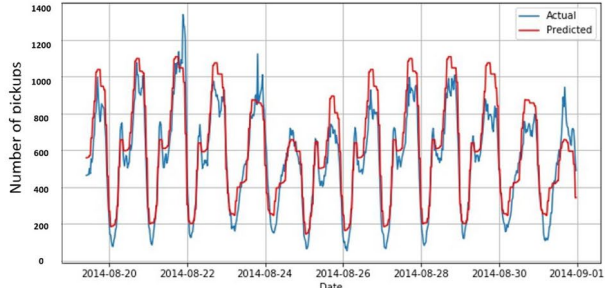
(j) UBERNET; 30min

◀ **Fig. 4** Performance of varying methods between 20/08/2014 and 01/09/2014 on the 15-min basis and 30-min basis, respectively (ARIMAX, PROPHET, LSTM, CNN, and UBERNET) importance in UBERNET by feature type and prediction hours of a day in the four boroughs. Feature importance is calculated as the mean absolute value of information gain (Theis and Bethge 2015). The importance of each feature set type is given by the maximum value of all features in that type. As expected, feature set A (temporal features) plays the most important part in all the four boroughs (since they provide direct short-term information about the pickups). For the short-term prediction problem, socioeconomic information, travel-to-work characteristics and social and built environment factors play a less important role. In addition, it is interesting to observe disparate patterns in the different boroughs. The model puts greater importance on temporal features in Manhattan and Queens, regardless of hours. In contrast, the model places less weight on temporal features in Brooklyn and the Bronx. Manhattan and the Bronx gains relatively higher importance on all features around 8:00 am to 14:00 pm, whereas in Queens and Brooklyn, higher importance is around 17:00 pm, when commuters return from work.

To shed some light on the spatial accuracy of UBERNET's advantage when compared with other models, Fig. 8d reports the performance comparison between UBERNET and CNN based on census data in Brooklyn. As can be seen, UBERNET outperforms CNN in the northern and the eastern part of Brooklyn, where the figure also shows some of the land-use and rich sociodemographic information employed in UBERNET, e.g., high crime rates area of Fig. 8a, high income area of Fig. 8b, and a variety of public transport area of Fig. 8c. The northern part of Brooklyn, where UBERNET outperformns the next best, CNN, the most, has the highest degree of transportation connectivity providing opportunities for pickup at transit stations and stops in a commercial area that also has higher crime levels on the average. This implies that the "cache" and the filter structure designed in

**Table 3** Performance of pickups prediction with different sets of features (RMSE)

| Learning model | Feature set [A] | Feature set [B] | Feature set [C] | Feature set [D] | All features |
|---|---|---|---|---|---|
| ARIMAX | 198.39 | 212.27 | 229.47 | 229.82 | 207.53 |
| LSTM | 189.34 | 207.70 | 235.81 | 227.04 | 179.28 |
| CNN-LSTM | 189.12 | **205.30** | 232.13 | 213.33 | 179.48 |
| SVR | 191.84 | 206.74 | 236.84 | 229.26 | 180.28 |
| Random forest | 192.16 | 207.82 | 230.71 | 226.25 | 180.60 |
| PROPHET | 223.43 | 225.89 | **227.35** | 215.27 | 198.63 |
| CNN | 192.36 | 211.64 | 229.39 | 229.21 | 183.97 |
| UBERNET | **188.42** | 215.73 | 228.37 | **211.17** | **177.55** |

The significance of bold value indicates that this model achieved the best performance

**Table 4** Ablation analysis of feature set

| Feature | RMSE | SMAPE | Feature | RMSE | SMAPE |
|---|---|---|---|---|---|
| pickups | 235.23 | 10.36 | professional: | 186.09 | 7.82 |
| hour: | 196.21 | 8.27 | wed: | 185.02 | 7.73 |
| day: | 195.17 | 8.19 | carpool: | 179.52 | 7.62 |
| borough: | 191.43 | 8.03 | totalPop: | 179.35 | 7.60 |
| income: | 189.96 | 7.97 | transp: | 179.87 | 7.55 |
| walk: | 187.52 | 7.89 | streetcrime: | 179.03 | 7.43 |
| umemployment: | 187.38 | 7.83 | hd: | 179.42 | 7.42 |
| vsb: | 187.15 | 7.83 | month: | 178.03 | 7.42 |
| pcp01: | 187.13 | 7.81 | PUMA: | 178.01 | 7.37 |
| pcp24: | 187.01 | 7.80 | meanCommute : | 178.01 | 7.37 |
| pcp06: | 186.92 | 7.80 | dewp: | 177.97 | 7.35 |
| temp: | 186.29 | 7.79 | workAtHome: | 177.95 | 7.33 |
| sd: | 186.35 | 7.78 | income: | 177.88 | 7.32 |
| spd: | 186.35 | 7.77 | slp: | 177.83 | 7.31 |

UberNet can better capture the complex interaction among the different features compared to the CNN.

### Time and Memory Cost

In this paper, all the experiments were conducted on a computer with XEON W-2125, 4.01 GHz processor and 64.00 GB RAM and two pieces of GTX 1080 Ti graphics card running on the system of Ubuntu 16.04.

Table 6 shows the computational cost and memory usage of varying models. As expected, ARIMAX, SVR, RF and Prophet have the least time and memory cost, due to their relative shallow structure and light weight implementation. LSTM consumes large amount of memory and time, since it requires additional memory to store the partial results for the calculation of multiple cell gate. However, in CNN-LSTM and UBERNET the filters are shared across a layer, where the backpropagation process only relies on the depth of the structure. We can also observe that UBERNET costs slightly more time and memory than CNN, which is probably due to the 1D dilated filters of UberNet whereas in CNN a standard 2D filter is used.

### Results Summary

From the point of view of transport operations, the main findings are as follows:

1. Compared to the other approaches, UberNet captures both the peaks as well as the troughs in ride-hailing demand, although it is more conservative compared to LSTM when predicting very steep surges.
2. UberNet achieved the best performance with respect to feature set A (temporal and real-time factors), which implies that the model is more responsive than others to pickup demand changes over time-of-day and day-of-week.
3. UberNet also performed better than other models when mining spatial features such as the census tract and its surrounding conditions such as crime levels and built environment factors i.e., the density of transport connection.

## Conclusions and Discussion

The motivation for this work is more accurate short-term forecasts of demand for Uber rides. The contribution of this paper is twofold. First, we presented UberNet, a deep learning architecture to address the problem of ride-hailing (Uber) pickup demand prediction. The results showed that UberNet outperforms the other learning models considered, including CNN and maintained its performance levels when used with smaller sample sizes in different boroughs of NYC.

A second contribution is the incorporation of insights from the transport research into machine learning problems for real-time transport predictions by incorporating various spatio-temporal features. The results have demonstrated that a multivariate architecture in deep learning, utilising each features, which have been suggested both by the transport operations as well as travel behavior research, has the possibility of significantly boosting model performance as well as insights into the nature of the Uber ride-hailing problem, in contrast to univariate deep learning approaches. As suggested by transport operations, introducing diurnal travel demand patterns, and as well as factors such as weather boosts prediction quality, which are insights not possible from univariate analysis. A key

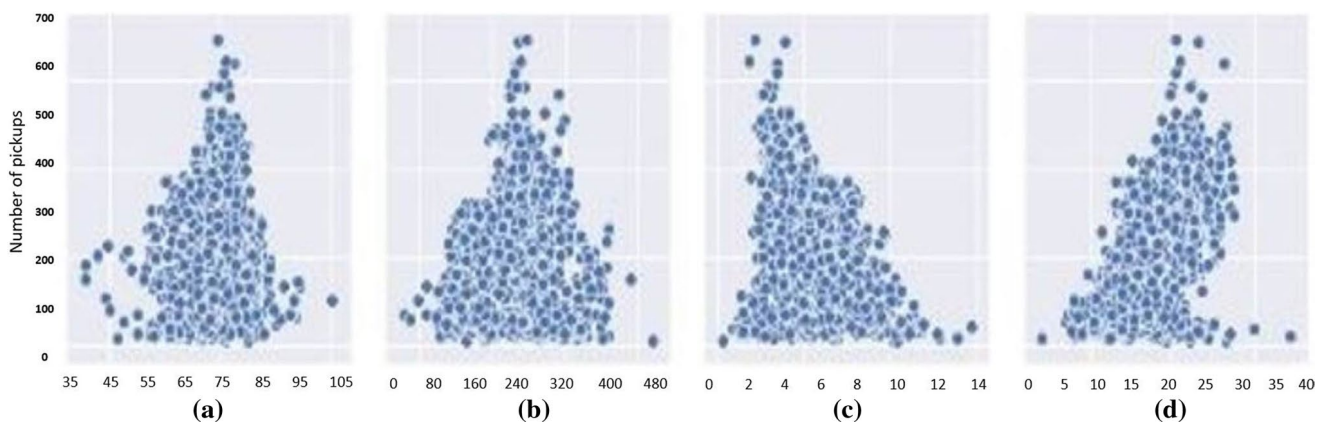

**Fig. 5** Partial dependence plots of **a** UberNet predicted pickups against temperature levels during pick-up time, **b** UberNet predicted pickups against street crimes rates, **c** UberNet predicted pickups against level of carpool commuters and **d** UberNet predicted pickups against density of transport facilities in pick-up area

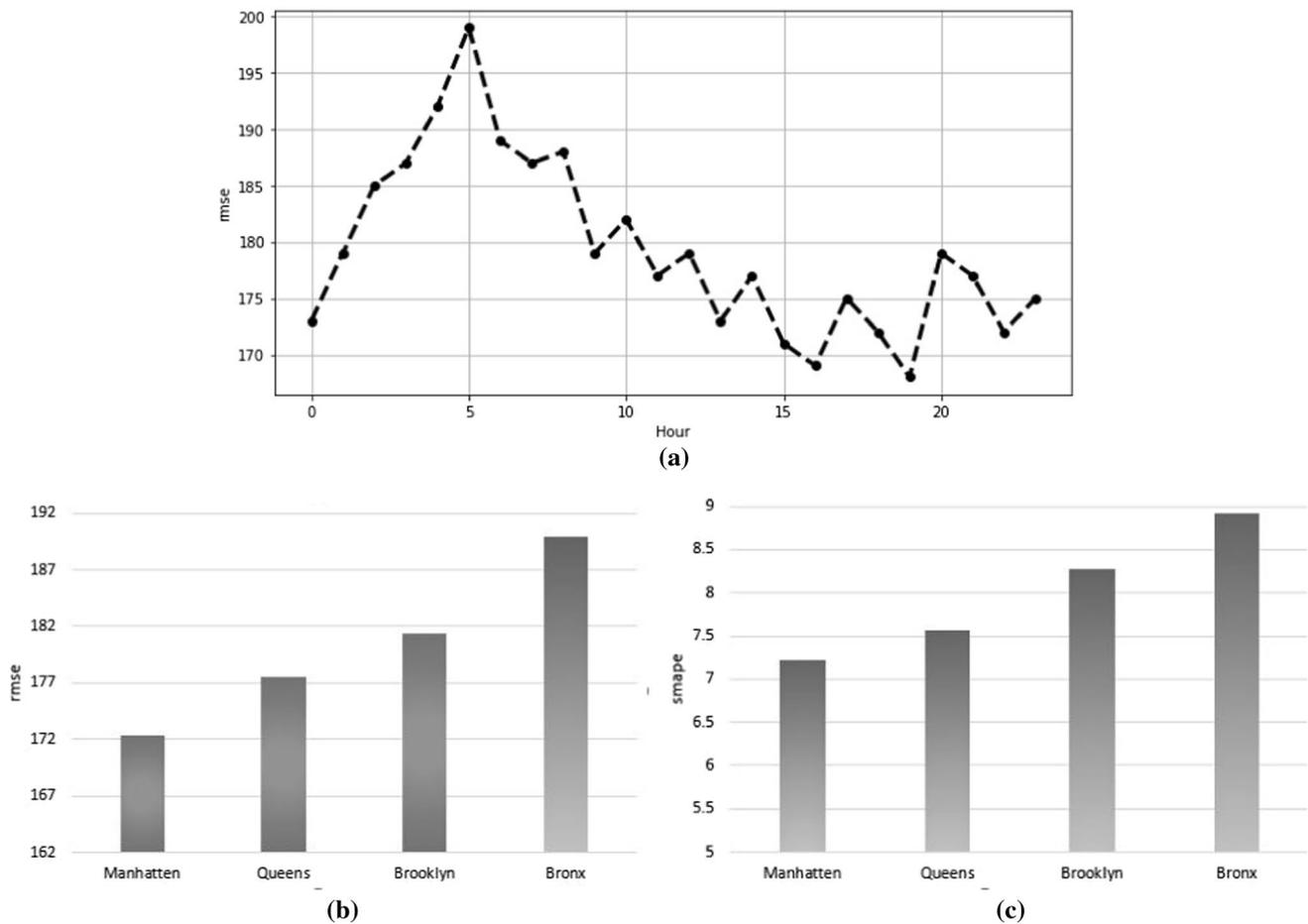

**Fig. 6** Comparison of UBERNET performance over: **a** hours of a day; **b** varying NYC boroughs with respect to RMSE, and; **c** varying NYC boroughs with respect to SMAPE

group of transport providers that are likely to benefit from accurate forecasts of increases in short-term demand for ride-hailing are public transport and demand-responsive transport providers. Through mechanisms such as connection protection and real-time dispatching, the connection between ride-hailing and mass transport and paratransit may become more seamless.

In addition, unique to this paper is the use of a number of socioeconomic, travel behavior and built environment characteristics within a deep learning framework for ride-hailing prediction. For instance, pickups increase as the level of crime increases, but only up to a point, after which pickups decline. This is potentially due to the interaction between crime rate and income levels, and the dependence on public transport rather than ride-hailing in low-income areas. These are issues that affect not only passenger demand, but also driver behavior. The analysis showed that the UBERNET approach can mine such information for increased accuracy in prediction. At the same time, the analysis identified the degrees of complexity of the relationships (e.g., non-linearity, bimodality) between such features and ride-hailing.

**Table 5** Performance of pickups prediction of varying models over Manhattan and Queens

| Borough | Metrics | ARIMAX | LSTM | CNN-LSTM | SVR | RF | Prophet | CNN | UBERNET |
|---|---|---|---|---|---|---|---|---|---|
| Manhattan | SMAPE | 8.19 | 7.26 | 7.34 | 7.98 | 8.20 | 8.17 | 7.43 | 7.18 |
|  | RMSE | 201.34 | 182.03 | 180.42 | 189.26 | 189.94 | 181.63 | 180.64 | 172.28 |
| Queens | SMAPE | 8.11 | 7.37 | 7.14 | 8.23 | 8.14 | 8.15 | 7.61 | 7.37 |
|  | RMSE | 203.58 | 179.42 | 178.07 | 182.72 | 183.60 | 192.83 | 181.25 | 171.89 |

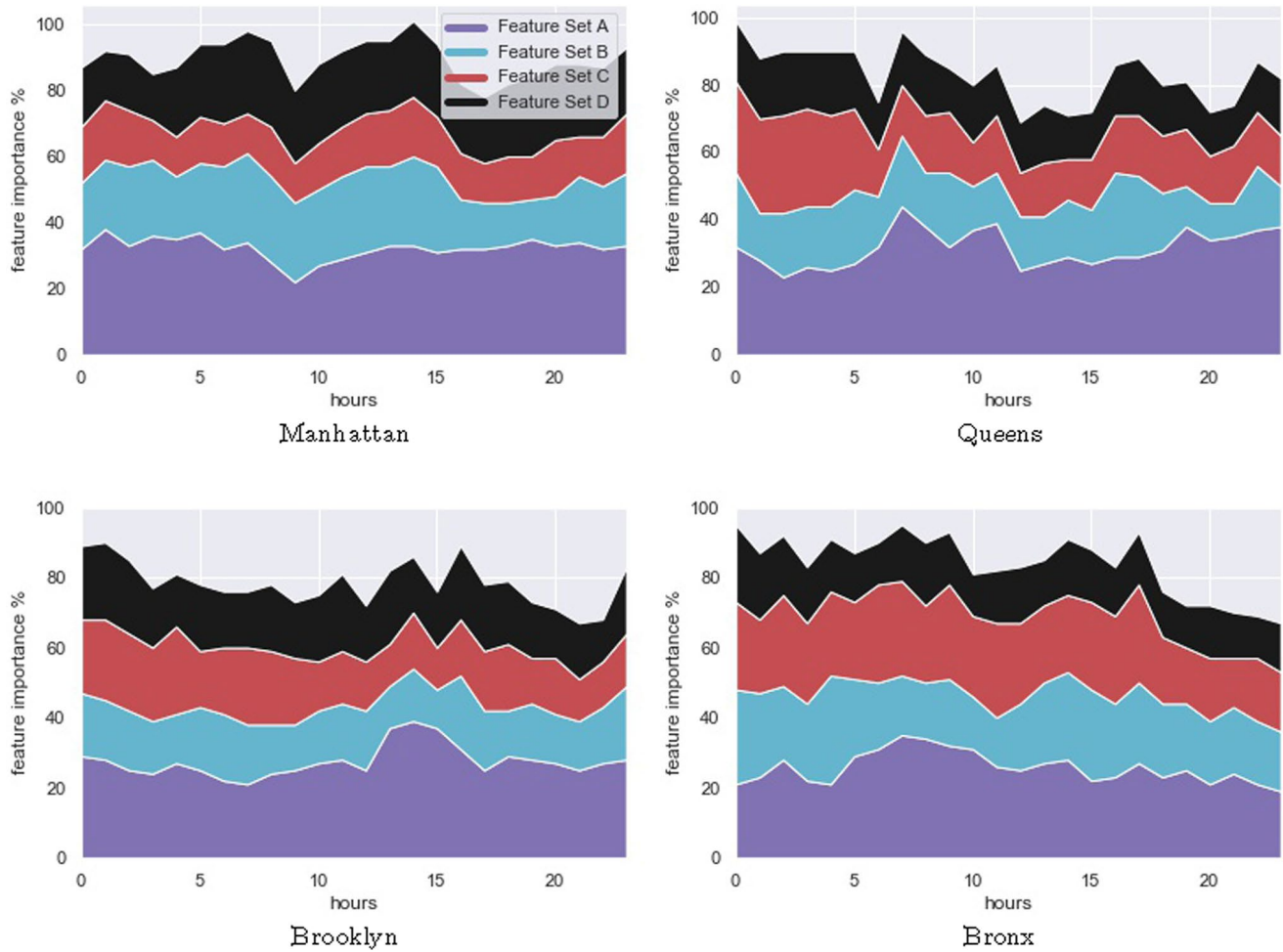

**Fig. 7** Relative feature importance in UberNet by feature set and prediction hours

We find that UBERNET was able to handle such complexities through its deep learning architecture.

There are several interesting and promising directions in which this work could be extended. First, users and Uber drivers can be naturally organized as a graph structure, and it will be interesting to explore the performance of some advanced graph embedding algorithms, such as BB-Graph (Asiler and Yazıcı 2017). Second, UBERNET in the current form is purely based on CNN structure, which makes sense as a first step towards integrating Uber's contextual environment into learning model, however, we could consider coupling it with more sophisticated structures, e.g., capturing regions with CNN features (R-CNN) (Girshick et al. 2015).

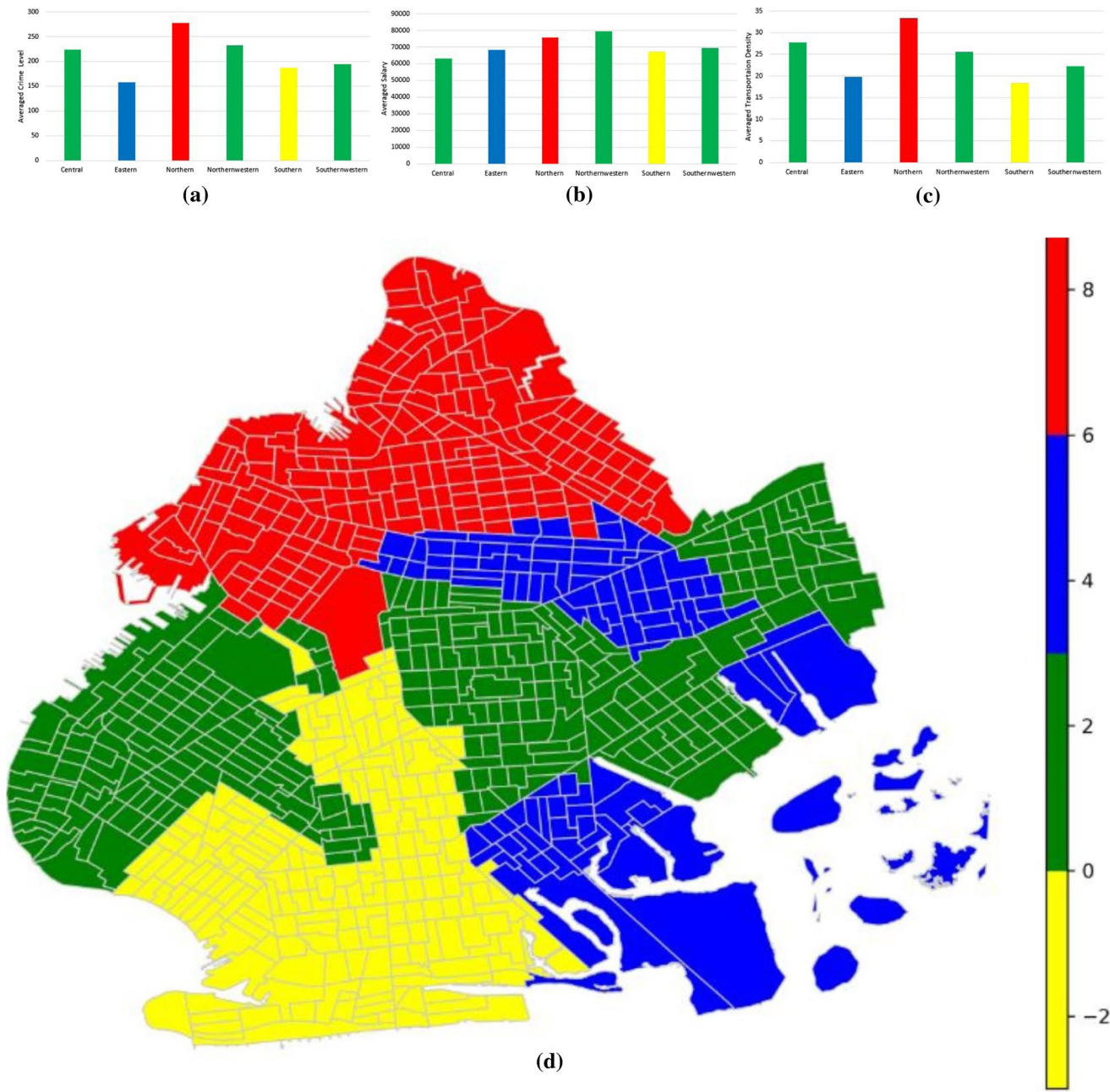

**Fig. 8** **a** Spatial heterogeneity of street crime level in Brooklyn; **b** spatial heterogeneity of income in Brooklyn; **c** spatial heterogeneity of transport density in Brooklyn; **d** Comparison of UberNet performance over *CNN* with respect to RMSE in the Brooklyn borough. Performance gain is defined as the difference between the RMSE of the UberNet and CNN

**Table 6** Computational cost and memory usage of varying models in the training stage

| Metrics | ARIMAX | LSTM | CNN-LSTM | SVM | RF | Prophet | CNN | UberNet |
|---|---|---|---|---|---|---|---|---|
| CPU-minutes | 1.82 | 97.91 | 66.23 | 3.68 | 8.37 | 3.27 | 47.35 | 49.72 |
| megabytes | 0.36 | 55.82 | 12.92 | 0.52 | 1.29 | 0.57 | 5.32 | 5.86 |

**Acknowledgements** This work was partially supported by the Economic and Social Research Council [grant number 301865-01].

**Open Access** This article is licensed under a Creative Commons Attribution 4.0 International License, which permits use, sharing, adaptation, distribution and reproduction in any medium or format, as long as you give appropriate credit to the original author(s) and the source, provide a link to the Creative Commons licence, and indicate if changes were made. The images or other third party material in this article are included in the article's Creative Commons licence, unless indicated otherwise in a credit line to the material. If material is not included in the article's Creative Commons licence and your intended use is not permitted by statutory regulation or exceeds the permitted use, you will need to obtain permission directly from the copyright holder. To view a copy of this licence, visit http://creativecommons.org/licenses/by/4.0/.

# References

Agiakloglou C, Newbold P (1992) Empirical evidence on dickey-fuller-type tests. J Time Ser Anal 13(6):471–483

Alemi F, Circella G, Handy S, Mokhtarian P (2018) What influences travelers to use uber? Exploring the factors affecting the adoption of on-demand ride services in California. Travel Behav Soc 13:88–104

Asiler M, Yazıcı A (2017) Bb-graph: a new subgraph isomorphism algorithm for efficiently querying big graph databases. arXiv:1706.06654

Battifarano M, Qian ZS (2019) Predicting real-time surge pricing of ride-sourcing companies. Transport Res Part C Emerg Technol 107:444–462

Bergstra J, Bengio Y (2012) Random search for hyper-parameter optimization. J Mach Learn Res 13:281–305

Bhat CR, Koppelman FS (1999) Activity-based modeling of travel demand. In: Hall R (ed) Handbook of transportation science. International series in operations research & management science, vol 23. Springer, Boston

Borovykh A, Bohte S, Oosterlee CW (2018) Dilated convolutional neural networks for time series forecasting. J Comput Finance **(forthcoming)**

Brodeur A, Nield K (2017) Has uber made it easier to get a ride in the rain? Tech. Rep. WORKING PAPER #1708E, University of Ottawa

Cervero R, Kockelman K (1997) Travel demand and the 3ds: density, diversity, and design. Transport Res Part D Transport Environ 2(3):199–219

Chen MK (2016) Dynamic pricing in a labor market: surge pricing and flexible work on the uber platform. In: Proceedings of the 2016 ACM conference on economics and computation. EC '16. ACM, New York, pp 455–455

Chen L, Mislove A, Wilson C (2015) Peeking beneath the hood of uber. In: Proceedings of the 2015 internet measurement conference. IMC '15. ACM, New York, pp 495–508

Chen D, Zhang Y, Gao L, Geng N, Li X (2017) The impact of rainfall on the temporal and spatial distribution of taxi passengers. PLoS One 12(9):e0183574

Cohen P, Hahn R, Hall J, Levitt S, Metcalfe R (2016) Using big data to estimate consumer surplus: the case of uber. Working Paper 22627, National Bureau of Economic Research

Cools M, Moons E, Wets G (2009) Investigating the variability in daily traffic counts through use of arimax and sarimax models: assessing the effect of holidays on two site locations. Transport Res Record 2136(1):57–66

Cramer J, Krueger AB (2016) Disruptive change in the taxi business: the case of uber. Am Econ Rev 106(5):177–82

Diego Correa KX, Ozbay K (2017) Exploring taxi and uber demand in New York city: empirical analysis and spatial modeling. In: Proceedings of the transportation research board 96th annual meeting

Drucker H, Burges Chris JC, Kaufman L, Smola A, Vapnik V et al (1997) Support vector regression machines. Adv Neural Inf Process Syst 9:155–161

Faghih S, Safikhani A, Moghimi B, Kamga C (2019) Predicting short-term uber demand in New York city using spatiotemporal modeling. J Comput Civ Eng 33(3):05019002

Gers FA, Eck D, Schmidhuber J (2002) Applying LSTM to time series predictable through time-window approaches. In: Neural nets WIRN Vietri-01, pp 193–200

Gerte R, Konduri KC, Eluru N (2018) Is there a limit to adoption of dynamic ride sharing systems? Evidence from analysis of uber demand data from New York city. Transport Res Record 2672(42):127–136

Girshick R, Donahue J, Darrell T, Malik J (2015) Region-based convolutional networks for accurate object detection and segmentation. IEEE Trans Pattern Anal Mach Intell 38(1):142–158

Goldstein A, Kapelner A, Bleich J, Pitkin E (2015) Peeking inside the black box: visualizing statistical learning with plots of individual conditional expectation. J Comput Graph Stat 24(1):44–65

Huang W, Song G, Hong H, Xie K (2014) Deep architecture for traffic flow prediction: deep belief networks with multitask learning. IEEE Trans Intell Transport Syst 15(5):2191–2201

Iglesias R, Rossi F, Wang K, Hallac D, Leskovec J, Pavone M (2018) Data-driven model predictive control of autonomous mobility-on-demand systems. In: 2018 IEEE international conference on robotics and automation (ICRA), pp 6019–6025

Kaji N, Kobayashi H (2017) Incremental skip-gram model with negative sampling. arXiv:1704.03956

Kamga C, Yazici MA, Singhal A (2015) Analysis of taxi demand and supply in New York city: implications of recent taxi regulations. Transport Plan Technol 38(6):601–625

Ke J, Zheng H, Yang H, Chen XM (2017) Short-term forecasting of passenger demand under on-demand ride services: a spatiotemporal deep learning approach. Transport Res Part C Emerg Technol 85:591–608

Kechyn G, Yu L, Zang Y, Kechyn S (2018) Sales forecasting using wavenet within the framework of the Kaggle competition. arXiv:1803.04037

Kim J, Kwan MP (2018) Beyond commuting: ignoring individuals' activity-travel patterns may lead to inaccurate assessments of their exposure to traffic congestion. Int J Environ Res Public Health

Lamb AM, Goyal AG, Zhang Y, Zhang S, Courville AC, Bengio Y (2016) Professor forcing: a new algorithm for training recurrent networks. In: Advances in neural information processing systems, pp 4601–4609

Laptev N, Yosinski J, Li LE, Smyl S (2017) Time-series extreme event forecasting with neural networks at uber. Int Conf Mach Learn 34:1–5

Li S, Tavafoghi H, Poolla K, Varaiya P (2019) Regulating TNCs: should Uber and Lyft set their own rules? Transport Res Part B Methodol 129:193–225

Liaw A, Wiener M et al (2002) Classification and regression by random forest. R news 2(3):18–22

Lin L, He Z, Peeta S (2018) Predicting station-level hourly demand in a large-scale bike-sharing network: a graph convolutional neural network approach. Transport Res Part C Emerg Technol 97:258–276

Lipton ZC, Kale DC, Elkan C, Wetzel R (2015) Learning to diagnose with LSTM recurrent neural networks. arXiv:1511.03677

Lv Y, Duan Y, Kang W, Li Z, Wang F-Y (2014) Traffic flow prediction with big data: a deep learning approach. IEEE Trans Intell Transport Syst 16(2):865–873

Meng M, Zhang J, Wong YD, Au PH (2016) Effect of weather conditions and weather forecast on cycling travel behavior in Singapore. Int J Sustain Transport 10(9):773–780

Neville J, Jensen D (2000) Iterative classification in relational data. In: Proc. AAAI-2000 workshop on learning statistical models from relational data, pp. 13–20

Nguyen HTA, Chikaraishi M, Fujiwara A, Zhang J (2017) Mediation effects of income on travel mode choice: analysis of short-distance trips based on path analysis with multiple discrete outcomes. Transport Res Record 2664(1):23–30

Niepert M, Ahmed M, Kutzkov K (2016) Learning convolutional neural networks for graphs. In: International conference on machine learning, pp 2014–2023

Polson NG, Sokolov VO (2017) Deep learning for short-term traffic flow prediction. Transport Res Part C Emerg Technol 79:1–17

Ramdas A, Tibshirani RJ (2016) Fast and flexible admm algorithms for trend filtering. J Comput Graph Stat 25(3):839–858

Rosenblat A, Stark L (2016) Algorithmic labor and information asymmetries: a case study of uber's drivers. Int J Commun 10:27

Ruder S (2016) An overview of gradient descent optimization algorithms. arXiv:1609.04747

Tan H, Xuan X, Wu Y, Zhong Z, Ran B (2016) A comparison of traffic flow prediction methods based on dbn. In: CICTP 2016, pp 273–283

Tang L, Thakuriah PV (2012) Ridership effects of real-time bus information system: a case study in the city of Chicago. Transport Res Part C Emerg Technol 22:146–161

Taylor SJ, Letham B (2018) Forecasting at scale. Am Stat 72(1):37–45

Thakuriah PV, Tilahun N (2013) Incorporating weather information into real-time speed estimates: comparison of alternative models. J Transport Eng 139(4):379–389

Theis L, Bethge M (2015) Generative image modeling using spatial LSTMs. In: Cortes C, Lawrence ND, Lee DD, Sugiyama M, Garnett R (eds) Advances in neural information processing systems, vol 28. Curran Associates, Inc., pp 1927–1935

Tilahun N, Thakuriah PV, Li M, Keita Y (2016) Transit use and the work commute: analyzing the role of last mile issues. J Transport Geogr 54:359–368

Tong Y, Chen Y, Zhou Z, Chen L, Wang J, Yang Q, Ye J, Lv W (2017) The simpler the better: a unified approach to predicting original taxi demands based on large-scale online platforms. In: Proceedings of the 23rd ACM SIGKDD international conference on knowledge discovery and data mining. ACM, pp 1653–1662

Van Den Oord A, Dieleman S, Zen H, Simonyan K, Vinyals O, Graves A, Kalchbrenner N, Senior AW, Kavukcuoglu K (2016a) Wavenet: a generative model for raw audio. In: SSW, p. 125 van den Oord A, Kalchbrenner N, Vinyals O, Espeholt L, Graves A, Kavukcuoglu K (2016b) Conditional image generation with pixelcnn decoders. CoRR. arXiv:1606.05328

Vlahogianni EI, Golias JC, Karlaftis MG (2004) Short-term traffic forecasting: overview of objectives and methods. Transport Rev 24(5):533–557

Wang H, Yang H (2019) Ridesourcing systems: a framework and review. Transport Res Part B Methodol 129:122–155

Wang D, Cao W, Li J, Ye J (2017) Deepsd: supply-demand prediction for online car-hailing services using deep neural networks. In: 2017 IEEE 33rd international conference on data engineering (ICDE). IEEE, pp 243–254

Wang C, Hao P, Wu G, Qi X, Barth M (2018) Predicting the number of uber pickups by deep learning. Tech. rep

Wang J, Chen R, He Z (2019) Traffic speed prediction for urban transportation network: a path based deep learning approach. Transport Res Part C Emerg Technol 100:372–385

Wu Y, Tan H, Qin L, Ran B, Jiang Z (2018) A hybrid deep learning based traffic flow prediction method and its understanding. Transport Res Part C Emerg Technol 90:166–180

Xu C, Ji J, Liu P (2018a) The station-free sharing bike demand forecasting with a deep learning approach and large-scale datasets. Transport Res Part C Emerg Technol 95:47–60

Xu Z, Li Z, Guan Q, Zhang D, Li Q, Nan J, Liu C, Bian W, Ye J (2018b) Large-scale order dispatch in on-demand ride-hailing platforms: a learning and planning approach. In: Proceedings of the 24th ACM SIGKDD international conference on knowledge discovery & data mining. ACM, pp 905–913

Yang S, Shi S, Hu X, Wang M (2015) Spatiotemporal context awareness for urban traffic modeling and prediction: sparse representation based variable selection. PLoS One 10(10):e0141223

Yang S, Ma W, Pi X, Qian S (2019) A deep learning approach to real-time parking occupancy prediction in spatio-termporal networks incorporating multiple spatio-temporal data sources. arXiv:1901.06758

Yao H, Wu F, Ke J, Tang X, Jia Y, Lu S, Gong P, Ye J, Li Z (2018) Deep multi-view spatial-temporal network for taxi demand prediction. In: Thirty-second AAAI conference on artificial intelligence

Yazdanbakhsh O, Dick S (2019) Multivariate time series classification using dilated convolutional neural network. arXiv:1905.01697

Young M, Farber S (2019) The who, why, and when of uber and other ride-hailing trips: an examination of a large sample household travel survey. Transport Res Part A Policy Pract 119:383–392

Yu X, Gao S, Hu X, Park H (2019) A Markov decision process approach to vacant taxi routing with e-hailing. Transport Res Part B Methodol 121:114–134

Zhang Y, Zhang Y (2018) Exploring the relationship between ridesharing and public transit use in the United States. Int J Environ Res Public Health 15:1763

Zhang J, Zheng Y, Qi D (2017) Deep spatio-temporal residual networks for citywide crowd flows prediction. In: AAAI, pp 1655–1661

Zhang K, Liu Z, Zheng L (2019) Short-term prediction of passenger demand in multi-zone level: temporal convolutional neural network with multi-task learning. In: IEEE transactions on intelligent transportation systems

Zoepf S, Chen S, Adu P, Pozo G (2018) The economics of RideHailing: driver expenses, income and taxes. Tech. rep., MIT Center for Energy and Environmental Policy Research